\pdfoutput=1
\documentclass[11pt]{article}

\usepackage[final]{acl}

\usepackage{times}
\usepackage{latexsym}
\usepackage{booktabs} 
\usepackage{setspace}

\usepackage{graphicx,color,framed,a4wide}
\usepackage[inline]{enumitem}
\usepackage[colorinlistoftodos]{todonotes}
\usepackage{arabtex}
\usepackage{utf8}
\setcode{utf8}
\definecolor{electriclime}{rgb}{0.8, 1.0, 0.0}
\definecolor{chartreuse}{rgb}{0.87, 1.0, 0.0}

\usepackage[T1]{fontenc}
\usepackage{microtype}

\usepackage{inconsolata}
\usepackage{utf8}
\usepackage{graphicx}
\usepackage{setspace}
\usepackage{xurl}
\usepackage{floatrow}
\usepackage[label font=bf,labelformat=simple]{subfig}% <-- changed
\usepackage{caption}
\newcommand{\majd}[1]{\textcolor{orange}{\bf\small [Majd: #1]}}

\title{Improving Language Models Trained on Translated Data \\with Continual Pre-Training and Dictionary Learning Analysis}

\author{\textbf{Sabri Boughorbel, Md Rizwan Parvez, Majd Hawasly}\\
Qatar Computing Research Institute \\
Hamad Bin Khalifa University, Doha, Qatar \\
\texttt{\{sboughorbel, mparvez, mhawasly\}@hbku.edu.qa} \\
}

\begin{document}
\setcode{utf8}
\maketitle
\begin{abstract}
Training LLMs for low-resource languages usually utilizes data augmentation from English using machine translation (MT). This, however,  brings a number of challenges to LLM training: there are large costs attached to translating and curating huge amounts of content with high-end machine translation solutions; the translated content carries over cultural biases; and if the translation is not faithful and accurate, data quality degrades causing issues in the trained model. In this work, we investigate the role of translation and synthetic data in training language models. We translate TinyStories, a dataset of 2.2M short stories for 3-4 year old children, from English to Arabic using the open NLLB-3B MT model. We train a number of story generation models of size 1M-33M parameters using this data. We identify a number of quality and task-specific issues in the resulting models. To rectify these issues, we further pre-train the models with a small dataset of synthesized high-quality Arabic stories generated by a capable LLM, representing 1\% of the original training data. We show, using GPT-4 as a judge and Dictionary Learning Analysis from mechanistic interpretability, that the suggested approach is a practical means to resolve some of the machine translation pitfalls. We illustrate the improvements through case studies of linguistic and cultural bias issues.

\end{list} % please keep as is 
\end{abstract}
\begin{figure}[!t]
        \centering
        \includegraphics[width=1\linewidth]{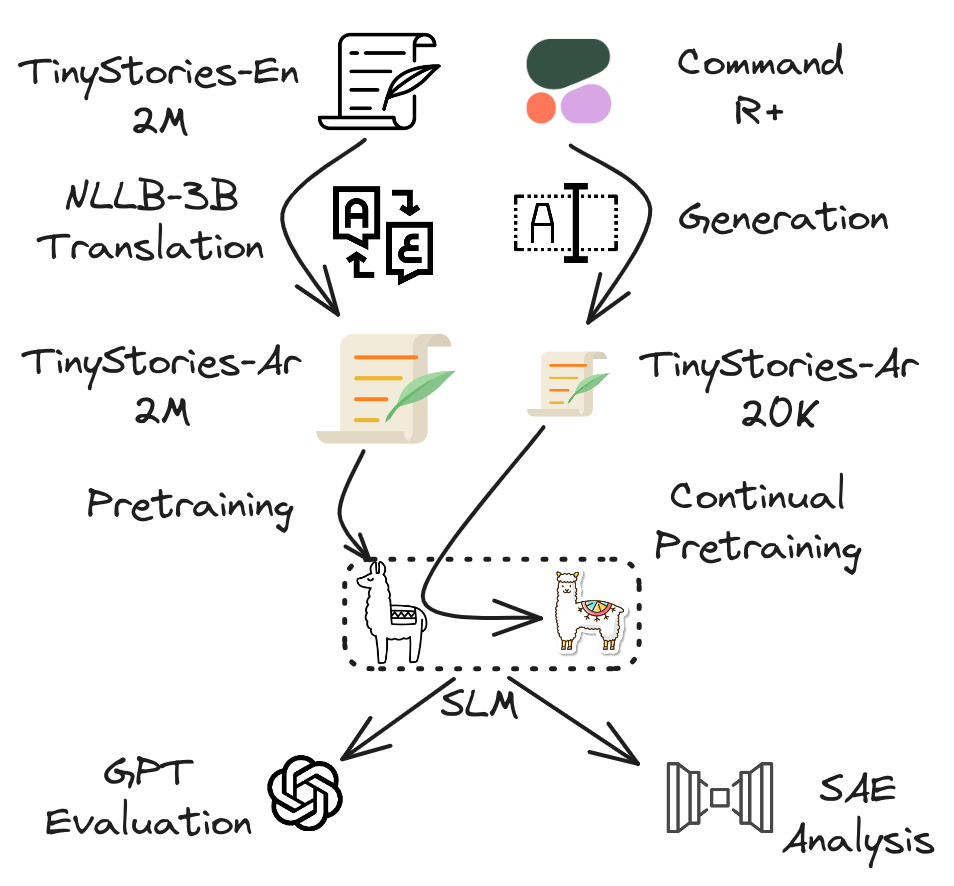}
        \caption{The proposed TinyStories Arabic dataset is formed by translating 2M tiny stories from English to Arabic using NLLB-3B and synthesizing 20K Arabic tiny stories using Command R+ LLM. The former data is used to pre-train small language models (SLMs) with different architectures. The latter is used for continual pre-training.  The models are qualitatively and quantitatively evaluated using a GPT judge. Further we train Sparse Auto-Encoder (SAE) on a selected SLM to analyze the model behavior.}
        \label{fig:enter-label}
\end{figure}

\section{Introduction}
Large Language Models (LLMs) have transformed the landscape of natural language processing (NLP) significantly. With the realization of unprecedented capabilities, the availability of underlying large training corpora and well-established language model training pipelines has shifted the focus in NLP research from defining linguistic inductive biases to the collection and curation of extensive text datasets \cite{dolma, penedo2024fineweb, together2023redpajama, openelm}. The recent trend showed an increase in focus on data curation and augmentation compared to innovation in model architecture or training paradigms \cite{brown2020language, touvron2023llama}. 
However, emergent capabilities in LLMs are noticeable for models of very large sizes such as 10B parameters or above and most of smaller LLMs of 200M-3B parameters have shown limited abilities in reasoning, fact recall and  coherent long  generation~\cite{schaeffer2024emergent}. While these models are typically trained with the same data used to train their larger and more capable siblings, the drops in capabilities of these small LLM are usually attributed to their reduced learnability or scale.  

However, newer small models have been recently shown to compete with 10x bigger models in challenging tasks. Examples include MiniMA~\cite{zhang2023towards} and the Phi family~\cite{abdin2024phi, li2023textbooks}. %A similar observation has been made for models with even smaller scale of tens of millions of parameters, such as the TinyStories models~\cite{eldan2023tinystories}.
These small language models (SLMs) offer faster training and lower deployment cost at the expense of being more task-specific and less general-use, which is not an issue for application-oriented language models. The key ingredient for the success of these smaller models seems to be the use of sophisticated and aggressive data curation, and high-quality synthetic data generated by  bigger LLMs. 

This new trend, however, does not map equally to languages that are not as privileged  with huge amounts of high quality content (or content in quantities that allow aggressive filtering), nor the availability of strong models that can be employed to generate diverse synthetic data in substantial quantities in a cost-effective manner, as is the situation with Arabic and many other low-resourced languages~\cite{thompson2024shocking}. 

A commonly-adopted workaround to the issue of data shortage is to turn to machine translation (MT) to benefit from the available content in English, which is evident in the data mixtures of the more capable Arabic models, e.g. Jais~\cite{sengupta2023jais}. However, the use of machine translation does not come without pitfalls. Specifically, 1) cultural biases that are stored within a language corpus get imported when translation is used, leading to misaligned models, e.g.~\cite{holmstrom-etal-2023-bridging}  and 2) based on the quality of the translation, certain linguistic intricacies of languages might not be respected~\cite{zhang-toral-2019-effect}, leading to degradation in the quality of the data and thus degradation in the capability of the final model with regard to the quality of the target language. 

In this work, we study this phenomenon using the recently-released TinyStories~\cite{eldan2023tinystories}, a synthetic dataset introduced to explore emergent properties in small language models. TinyStories comprises 2.2M short stories in English of about 200 words generated by GPT-3.5 and GPT-4. The relatively small models trained on TinyStories have shown interesting capabilities in generating coherent and creative short stories with correct grammar.  %Thus, TinyStories dataset and models are an excellent testbed for a better understanding of emergent behaviors in large language models. 
We translate TinyStories to Arabic using the open-source translation model NLLB-3B to simulate medium-quality translation. We train models with different sizes using the  Arabic-translated TinyStories, and benchmark the trained models against several other Arabic LLMs using GPT-4 as a judge following previous works \cite{zheng2024judging} in the task of story generation using three metrics: grammar correctness, consistency with the provided context, and creativity. We also identify some cultural and linguistic issues that arise from the translation. Then, we synthesize a high quality small dataset of 20K stories to explore the efficacy of continual pre-training in recovering from the issues brought along by low-quality translated data.

The contributions of this work are as follow: 
\begin{itemize}
\item We investigate the degradation of language models when using translated data of medium quality in training. We identify linguistic issues and cultural biases, which we address by  further pre-training the models with a limited amount of high-quality synthetic data.% We benchmark different architectures of small language models. 
\item We create a dataset with an Arabic version of TinyStories through medium quality translation, plus a small high-quality synthetic data which we will release as open-source to facilitate studying translation issues.
\item  We provide a comparative analysis before and after refinement using dictionary learning methods from mechanistic interpretability to assess the effects of continual pre-training with high-quality data. 
\end{itemize}

The rest of the paper is organized as follows: First, we describe how we prepare training data. Then we discuss the limitations of translated data. In Section~\ref{sec:continue_pre-training}, we show how continual pre-training with small amount of high-quality improve the models. 
% \majd{The contributions of this work are as follows:
% }
% \majd{The rest of the paper is organized as follows ...}

%For SLM training, we first validated our training pipeline using TinyStories English. 
%Despite the rising interest in developing Arabic LLMs, existing models shows some limitations in grammar, poetry and other aspects. The mainstream web data in Arabic language have several limitations: Limited scientific and STEM related topics are written about. Knowledge about Arabic grammar and stylistic writing is becoming scarce and hence this is also reflected in web data. Alternatively synthetic data and translation needs to explored to fill these knowledge gaps. 

%Exploring the role of synthetic and high quality data for training small Arabic language models is a promising direction. The requirement in terms of data volume and compute resource is affordable for a wide community  of Arabic NLP. The capability of small language models when trained on specific task and high quality data is surprising good. 

\section{Story Generation SLMs}
\subsection{TinyStories dataset}
First we give a brief description of the original TinyStories dataset \cite{eldan2023tinystories}. It consists of 2.2M short stories for 3-4 year old kids generated by GPT-3.5 and GPT-4. A set of 1500 basic words consisting of nouns, verbs and adjectives are initially collected that can be easily understood by a 3-year-old child. For each story, a verb, a noun and an adjective are randomly selected from the list. GPT models are prompted to generate a story including all the selected words, and that respects certain storytelling features, like a certain plot or a way of ending
as could be seen in Table~\ref{tab:gen_features}  in the Appendix. 

\subsection{Translated TinyStories}
We translated the  dataset using an open source machine translation  model that belongs to the `No Language Left Behind' project~\cite{costa2022no} which is intended primarily for 200 low-resource languages, offering a family of five models with sizes ranging from  600M parameters to 54B parameters. NLLB has good Arabic translation performance  compared to other solutions~\cite{tiedemann2023democratizing, kudugunta2024madlad}. %Mainly we had three options to consider for the automated machine translation: 1) MT models via API calls: There are free and paid API-based models. In both cases the cost and time required to translate the full dataset are very high. \majd{2), 3) ...}
We chose NLLB-3B, a model that can be deployed locally with reasonable hardware requirements, e.g, a single GPU with 16GB memory. The 3B size was chosen with practicality in mind, and to simulate the use of medium quality translation when preparing data on a large scale for language model training.

\subsection{Evaluation Method}

We followed a similar evaluation approach to the previous work on TinyStories~\cite{eldan2023tinystories}. We translate the test set of 44 manually-picked story completion prompts from English to Arabic using GPT-4. An example story completion  prompt from the test set could be seen in Appendix~\ref{appx:synth}. %the top panel of Figure~\ref{fig:dd}. %The English prompts were prepared manually as story beginnings. 
The choice of GPT-4 as a translator was to obtain a high-quality test set, and to ensure that the test data is out-of-distribution to the training data from the translation perspective. %  The test set consists of 44 prompts starting stories. The last part of each test prompt is an incomplete sentence. The models need to continue the story completion from there. 

During evaluation, GPT-4 is also used as a judge to assess the story completion of the tested models. The LLM judge is asked to pay extra attention to the initial sentence that is cut short, and to assess three main qualities: correctness of grammar, creativity of the story, and consistency to the prompt and the story details. Each of the metrics is given an integer score between 0 and 10. For each model, two completions with temperature 1 are generated for each test prompt, then average  scores are reported. 

\subsection{Model Training}
We use the translated dataset to train a number of small language models. We consider the same model size and configuration as in the original TinyStories family\footnote{\url{https://huggingface.co/roneneldan}}. The only difference was in replacing GPT-NeoX architecture  with Llama-2. This choice was based on the success and wider adoption of Llama to train LLMs. The main contrast between the two architectures  is that LLama-2 uses gated MLP layers, grouped-query attention and an RMS normalization, whereas GPT-NeoX uses parallel attention MLP and Layer-Norm normalization \cite{zhang2019root, liu2021pay, ainslie2023gqa, black2022gpt}. In the chosen model family, the number of heads is fixed, whereas the number of layers and residual stream (latent space) are varied. In 1M, 3M and 8M, the number of layers is kept constant at 8, while the dimension of residual stream is increased. In the remaining models, the increase in the number of layers is compensated by a reduction in the residual stream dimension. The goal is to maintain the model size in the range of 20M-30M parameters. Table~\ref{tab:architecture} provides a summary of the architecture of trained models. %Model dimensions are chosen same as in TinyStories paper for the sake of comparison.

\begin{table}[!ht]
    \centering
\begin{tabular}{l|ccc}
          \toprule & \small Layers & \small Heads & \small Hidden size \\ \toprule
\small \textbf{33M-ar} & 4        &   16       &  768     \\
\small \textbf{28M-ar} & 8         & 16       & 512      \\
\small \textbf{2L-33M-ar}        &    2       &    16      &   1024       \\
\small \textbf{1L-21M-ar}        &   1        &     16     &     1024     \\
\small \textbf{8M-ar }       & 8         & 16       & 256      \\
\small \textbf{3M-ar}        & 8         & 16       & 128      \\
\small \textbf{1M-ar}        & 8         & 16       & 64      \\ \bottomrule
\end{tabular}
\caption{Architecture of trained models %architecture. L, H and D denotes the numbers of layers, heads and hidden size respectively.
} \label{tab:architecture}
\end{table}

For SLM training, the choice of tokenizer is of utmost importance. Since the embedding layer can consume a large proportion of trainable parameters, it is crucial to carefully choose the vocabulary size and tokens. Existing tokenizers suited for Arabic language models usually include English or are over-sized for our purpose of training SLMs. We decide to train a byte-pair encoding (BPE) tokenizer on the translated training data. We experimented with different vocabulary sizes (8k, 16k, 32k and 85k). The  last two choices are motivated by the Llama-2 and Jais tokenizers. We trained a reference model (33M) with the different tokenizers to evaluate the role of vocabulary size on performance. As shown in Table~\ref{tab:vocab_size}, we selected a vocabulary size of 32k given it performed the highest.

\begin{table}[h!]
\centering
\begin{tabular}{l|ccc} \toprule
\small{Vocab size} &  \small{Grammar} & \small{Creativity} & \small{Consistency} \\ \midrule
\small\textbf{8k} & 2.97 & 3.97& 2.69\\
\small\textbf{16k} &4.64 &  4.72&  4.21\\
\small\textbf{32k} & \textbf{6.83}& \textbf{7.01} & \textbf{7.20}\\
\small\textbf{85k} & 3.02& 3.54&  1.77\\
\bottomrule
\end{tabular}
\caption{Evaluation of different vocabulary sizes on the performance of a selected SLM.}
\label{tab:vocab_size}
\end{table}

 For model training, we used the English data and published benchmarks as a means to calibrate the training process, see Appendix~\ref{appx:calibration}. We use AdamW optimizer with constant learning rate of $5\,.\,10^{-5}$ as in TinyStories paper with 5\% steps for warm-up~\cite{eldan2023tinystories}.  %
 %We used a learning rate of \majd{??}.

%Figures~\ref{fig:loss_train_ar} and \ref{fig:loss_eval_ar} show the training and validation loss for the trained models, respectively. 
The training and validation loss of the training can be seen in Appendix~\ref{appx:pt_loss}. The shape of the loss curves indicate that the training settings are adequate. Models with lower loss have better benchmarking performance as we will discuss later. The convergence loss for models trained on Arabic data are higher than the models trained on English data. One justification could be that the noise introduced by translation makes the task of next token prediction more difficult than the original task.

% \subsection{Results}
%We validated our training scripts by training the SLMs on the original English TinyStories. Also we validated our evaluation pipeline by comparing the SLMs with a few English LLMs.

%The architecture of trained models are similar to English TinyStories ones. Those choices help exploring the role of model depth and width on the performance.
Table~\ref{tab:model_dim} shows the GPT-4 scores on the three benchmarking dimensions (grammar, creativity and consistency) for the trained models on translated TinyStories. The best performance is achieved between the 28M and 33M models, and we will use the latter for the rest of this paper. An example completion of the 33M-ar model could be found in Appendix~\ref{appx:synth}.%could be seen in the lower panel of Figure~\ref{fig:dd}. %which is \majd{not the largest but has the largest depth and hidden size combination.}

\begin{table}
\centering
\begin{tabular}{l|ccc} \toprule
\small{Model} & \small{Grammar} & \small{Creativity} & \small{Consistency} \\ \midrule
\small{\textbf{28M-ar}} &6.80 & \textbf{7.32}& \textbf{7.28}\\
\small{\textbf{33M-ar}} &\textbf{6.83} & 7.01& 7.20\\
\small{\textbf{2L-33M-ar}} &6.52 & 7.01&  7.00\\
\small{\textbf{1L-21M-ar}} &5.87 & 6.11& 5.96\\
\small{\textbf{8M-ar}} & 6.47& 6.84&  6.70\\
\small{\textbf{3M-ar}} &5.82 & 6.48& 5.79\\
\small{\textbf{1M-ar}} &4.37 &  5.11& 3.43\\
\bottomrule
\end{tabular}
\caption{Benchmarking the trained models using GPT-4 as a judge on grammar, creativity and consistency}
\label{tab:model_dim}
\end{table}
We also compare the results of our trained SLMs with some state-of-the-art Arabic or multilingual   LLMs of sizes 200M-13B~\cite{radford2019language, antoun-etal-2021-aragpt2,koubaa2024arabiangpt,workshop2023bloom,kamal-eddine-etal-2022-arabart,elmadany-etal-2023-octopus,sengupta2023jais} in Table~\ref{tab:model_ar}.

\begin{table}[h!]
\centering
\begin{tabular}{l|ccc} \toprule
\small{Model} & \small{Grammar} & \small{Creativity} & \small{Consistency} \\ \midrule
\small{\textbf{gpt2-small}} &3.23 & 2.43& 1.38\\
\small{\textbf{aragpt2}} & 3.02& 2.84& 1.76\\
\small{\textbf{arabian-gpt}} &3.86 &4.27 &2.65 \\
\small{\textbf{bloom-1b1}} & 5.34& 4.75& 4.00\\
\small{\textbf{AraBart}}  &7.07 &5.03 & 5.96\\\midrule
\small{\textbf{33M-ar}} &6.83 & \textbf{7.01}& {7.20}\\ \midrule
\small{\textbf{AraT5v2}} &8.03 &6.69 &8.33 \\
\small{\textbf{jais-13b}} & \textbf{8.10}&6.89 & \textbf{8.43}\\
\bottomrule
\end{tabular}
\caption{Comparing the 33M-ar model to some state-of-the-art Arabic models using GPT-4 as a judge}
\label{tab:model_ar}
\end{table}

\section{Navigating Pitfalls of Translated Data}
 There are issues faced when we train models with medium-quality translation: 1) \textbf{Cultural biases}: translated data carries over the culture of the source data into the target language, which might not be desirable. We choose in this work one example from this category to focus on, which is that the translated stories come with English person names, leading to an Arabic model that can only generate stories with such names; 2) \textbf{Grammatical and style issues}: languages differ in how they express the same sentence, and weaker translations might fail to correct for nuanced style issues. We choose a single example related to how the speaker could follow the saying in English but not in Arabic. For example, the direct Arabic translation of the English sentence ``\texttt{`I'm happy!', said Tim.}'' to \begin{arabtext}{\small `أنا سعيد!'،
 قال تيم.} \end{arabtext} is not appropriate.%;  3) Translated data contain traces of English text due to noisy text in the original dataset. For example, we noticed the presence of emojis and symbols in the original TinyStories. Filtering stage on source data is needed to reduce the contamination of the translated data with such noise.  We plan to include such filtering in future work. %\majd{?}.

% In order to validate these qualitative observations, we turn to mechanistic interpretability tools that allow us to gain holistic insights into the trained models.

\subsection{Model refinement with continual pre-training}
\label{sec:continue_pre-training}
In order to tackle some of the MT  issues, we test whether the trained models can be improved with a second stage of training  utilizing a small amount of high-quality data that does not have the aforementioned issues.  This practical and cost-effective approach is somewhat similar to curriculum learning where LLMs are exposed to different levels of complexity at different stages of training \cite{zhou2020curriculum, chang2021does}.

\subsubsection{Synthetic Generation}
\label{ssec:synth}
We synthesized an additional small amount of high-quality stories using an open LLM capable in Arabic. %The two open LLMs that we selected were Llama-3 (30B) and Command R+ (100B). We performed qualitative assessment via prompting to select the better model.
We chose Command R+ (104B) for the task as it is the more capable model in Cohere's family of multilingual models which scores highly on Huggingface's Open Arabic LLM Leaderboard~\cite{OALL}. This new high quality content corresponds to 1\% of the original noisy training data. Instead of deploying Command R+ locally (high compute requirement of $3\times$ A100-80GB GPUs for inference), we  generate the 20K stories via paid API calls for a cost of \$100. %This high quality data is kept for a continual training of SLMs pre-trained on the translated data.
% \subsection{Synthesizing High Quality Stories}
% 
%In order to address the issues of medium-quality translation data for training the language models, we generate additional data using a strong LLM in Arabic. One of the open models available is  Command R+.  It is currently ranking high in English benchmarks and has been reported to have good performance on Arabic. Our preliminary testing with prompting showed that command R+ is a good candidate for synthesizing data. 
We populated the prompts with samples from a set of random Arabic verbs, nouns and adjectives suited for 3-year-old stories, in addition to a selection story features as used in the original dataset. %, see Table~\ref{tab:gen_features} in the Appendix. The prompt template is shown in Figure~\ref{fig:prompt1}.
More details of the process can be found in Appendix~\ref{appx:synth}.

\subsubsection{Continual pre-training}
We further pre-train the 33M-ar SLM on the high-quality synthetically-generated data from the optimizer state saved after a single full epoch of training on the translated data. The additional data represent about 1\% of the original training data. Therefore, the additional training time is small. The loss curves for the continual pre-training could be seen in Appendix~\ref{appx:cpt_loss}.%Figures~\ref{fig:loss_train_ar_fpt} and \ref{fig:loss_eval_ar_fpt} show the training and validation loss for the continual pre-trained 33M model, respectively. 
The benchmarking results after refinement can be seen in Table~\ref{tab:continue_pre-train}. Continual pre-training shows performance improvement over the three metrics. Other refinement strategy such as multi-epochs, mixing with translation data in pre-training and then further pre-training with high-quality data will be explored in future work.
\begin{table}
\centering
\begin{tabular}{l|ccc} \toprule
\small{Models} &  \small{Grammar} & \small{Creativity} & \small{Consistency} \\ \midrule
\small{\textbf{28M-ar}} &6.80 & {7.32}& 7.28\\ 
\small\textbf{33M-ar} & 6.83& 7.01 & 7.20\\ \hline
\small\textbf{33M-ar-CP} & {7.08} & {7.21}&  {7.26}\\
\bottomrule
\end{tabular}
\caption{Performance comparison of the continued pre-trained model (33M-ar-CP) and the best two base models trained only on translation data (28M-ar and 33M-ar). We choose the largest model as baseline in this experiment.}
\label{tab:continue_pre-train}
\end{table}
% \section{Mechanistic Interpretability Tools}
\subsection{Interpretability using Sparse Auto-Encoders}
In order to investigate whether the refinement process has managed to deal with the identified issues, we turn towards mechanistic interpretability. 
The traditional neuron-based methods have been shown recently not to be sufficient as neurons in language models appear to be \textit{polysemantic}~\cite{bricken2023towards}, i.e., a single neuron might activate on multiple unrelated concepts, and traditional interpretability approaches cannot disentangle the real cause of a certain phenomenon. Recently, Dictionary Learning~\cite{elhage2022superposition} has been developed as an alternative tool to interpret language models. The use of Sparse Auto-Encoders  (SAE)~\cite{bricken2023towards, rajamanoharan2024improving, makelov2024towards} to project neuron activations to a sparse and large dimensional \textit{feature} space has led to a highly-interpretable representation that captures knowledge in generative language models not seen at neuron levels. 

In order to understand the effect of our continual pre-training, we train a Sparse Auto-Encoder for both  base and continually-trained models on the output of the last MLP layer. For these analyses we choose 2L-33M-ar model\footnote{This model has previously been used for SAE analysis on English TinyStories:~\url{https://huggingface.co/collections/lovish/tinystories-sae-regularization-comparison-65edb87a85cb3a3386c4b062}} (see Table~\ref{tab:model_dim}). 

\subsubsection{Training SAEs}
The architecture of our SAE is one hidden layer MLP trained as an autoencoder, with input weights as an encoder and output weights as the decoder. We chose an expansion factor of 16. The training data for SAEs is prepared as follow: For each context in the training data, the MLP activation vectors are collected after the SILU non-linearity for each token in the context. Activation vectors are sampled from 128 tokens within each context. These sampled vectors are then shuffled together, ensuring that the samples in a given batch originate from a variety of different contexts. Adam optimizer is used to minimize an objective function consisting of two components: the mean squared error loss and an L1 regularization term. The L1 regularization penalty is applied to the activations of the hidden layer within the autoencoder architecture. For a given feature the logit weight is defined as the product of a feature direction and the unembed $W_U W_{dec}[feature]$. The logit weight measures the direct effect of the features on the likelihood of next-token prediction.  Examples of logit weight distribution can be found in~\cite{lin2024announcingneuropedia}\footnote{\url{https://www.neuronpedia.org/gpt2-small/8-res-jb/6649}}.

\subsubsection{Token Set Enrichment Analysis}
The Token Set Enrichment Analysis (TSEA) introduced in \cite{bloom2024understandingfeatureslogitlens} is borrowed from Gene Set Enrichment Analysis in bioinformatics~\cite{subramanian2005gene} to statistically quantify if a set of features \iffalse \majd{features?} \fi is strongly associated with a hypothesis. TSEA examines the \iffalse \majd{this sentence need revision:} \fi relationship between the distribution of logit weights and predefined sets of tokens. The latter could represent various semantic or linguistic categories such as set of token of boy or girl names. TSAE would in that case identify which features inhibit or promote these sets of names. \iffalse \majd{|} \fi The steps in TSEA  are: 1. Generate a library of token sets for the hypothesis under investigation. 2. Compute the \textit{enrichment scores} for all features across the sets. The enrichment scores are running sum statistics that expose which features promote or suppress tokens in the hypothesis sets. 3. Identify elevated points in the enrichment scores. 4. Inspect features with high enrichment scores to validate the hypothesis. 

\subsubsection{Analyzing MT issues with TSEA}
\label{sec:mt_issues}
In this work, we investigate two hypotheses as case studies of the effect of continual pre-training with high-quality data: 1. The cultural bias of carrying English names to translated data. 2. The English language dialog follows the template \texttt{``[Quoted text],'' said [Person]}, while in Arabic, the template is  \texttt{Said [Person]: ``[Quoted text]''}. 

\begin{figure}[!htb]\centering
    \includegraphics[width=1\textwidth]{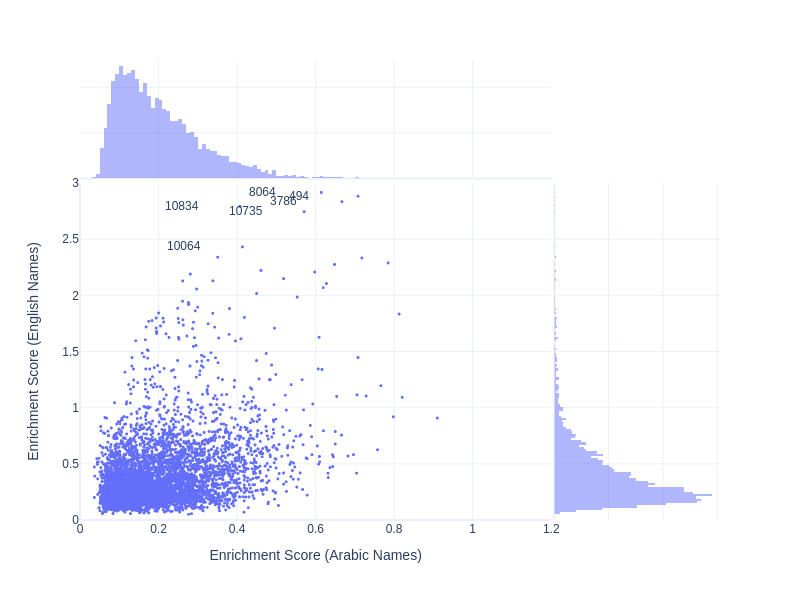}
    \caption{Scatter plot of feature enrichment scores for Arabic and English names in  2L-33M-ar.}
    \label{fig:tsea_names_translation}
\end{figure}

\begin{figure}[h!]
    \centering
    \includegraphics[width=1\textwidth]{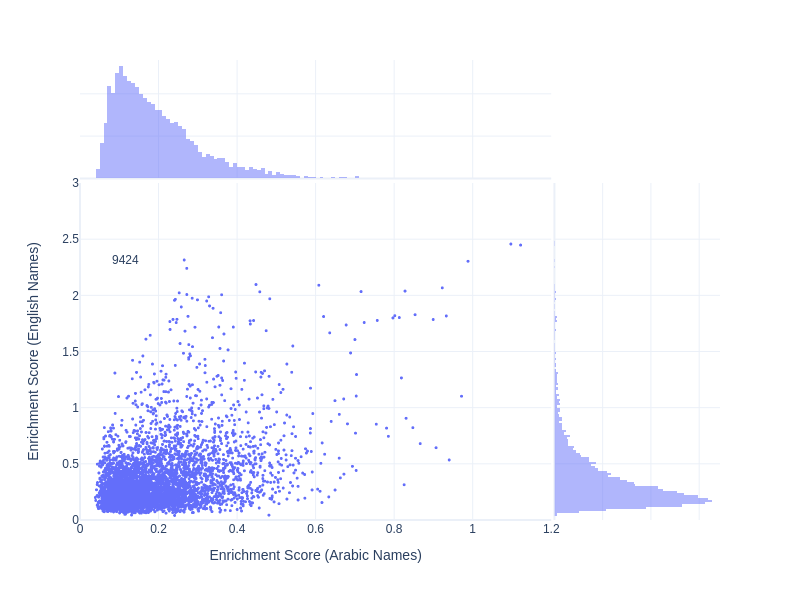}
    \caption{Scatter plot of feature enrichment scores for Arabic and English names for 2L-33M-ar-CP model after further pre-training.}
    \label{fig:tsea_names_continual_pre-training}
\end{figure}

For cultural bias, we defined two token sets of 600 common English and Arabic first names. TSEA help identifying which set of names is higher represented in the model. Figure~\ref{fig:tsea_names_translation} shows the scatter plot of enrichment scores for English vs. Arabic names. Features above the diagonal activate stronger on English  than Arabic names, and vice versa. We identified and marked the features farther from the diagonal with at least 2 point difference between the English and Arabic name enrichment scores, indication strong bias. The plot shows a higher concentration of features representing English names, while the similar plot after continual pre-training in Figure~\ref{fig:tsea_names_continual_pre-training} shows only one feature has the gap. The overall distribution is corrected away from English names, as also could be seen in generated samples. This illustrates the benefit of continual pre-training in correcting cultural bias.

\begin{figure*}[!htb]
    \centering
    \includegraphics[width=0.94\linewidth]{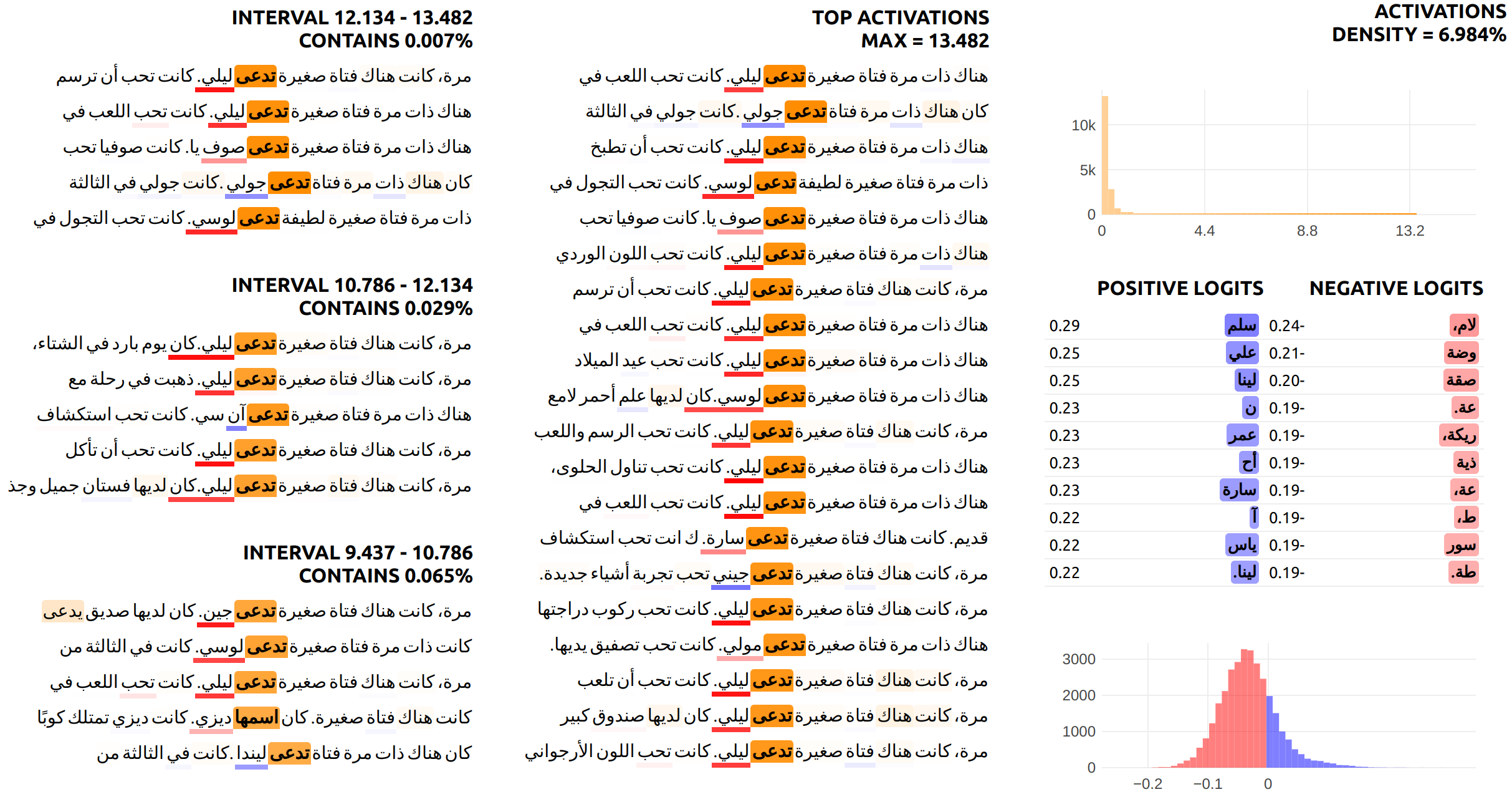}
    \caption{Dashboard of feature \#3114 from SAEs trained on the last MLP layer of continually trained model 2L-33M-ar-CP.  The Feature \#3114 shows that cultural bias was corrected after continual pre-training.}
    \label{fig:dashboard_names_dot}
\end{figure*}

The second case study is on the linguistic issue related to inappropriate dialog tagging in Arabic, which we noticed to be frequent in the translated TinyStories due to the quality of NLLB-3B. We defined the token set with English and Arabic persons names ending a sentence, and  performed TSEA analysis before and after continual pre-training. Figure~\ref{fig:dashboard_names_dot} shows the Manhattan plots of enrichment scores of both case studies. The first panel on the left shows the scores for dialog tagging for the model trained on translation data (a) and after continual pre-training (b). We carefully inspected the dashboards of top-3 features (2322, 3353, 14589) and (50, 1455, 9578). We concluded \iffalse \majd{where?} \fi that the top-features in the updated model activate on correct dialog tag phrases whereas the base model shows clearly the issue. 

Feature \#3144 is a good example to illustrate both case studies. Figure~\ref{fig:dashboard_names_dot} shows the feature dashboard\footnote{A guide to read the dashboard could be found in~\url{https://transformer-circuits.pub/2023/monosemantic-features/index.html\#setup-interface} and more details can be found in \cite{bricken2023towards, bloom2024gpt2residualsaes}}. The histogram on the top right is of sampled nonzero activations. The list of token of positive and negative logits underneath are the lowest and highest logit difference tokens of that feature, i.e., tokens whose probabilities of being sampled decrease or increase the most when the feature activates. Notice that the highest ranked tokens are of Arabic names, meaning that the model favors to produce an Arabic name after the verb \begin{arabtext}{\small تدعى} \end{arabtext} (which means "called") that activates the feature highly as seen in the top activation panel to the left. The color highlights activation level of the token. A blue underline represents a lower loss (better prediction of that token), whereas a red underline represents a higher loss. The top activation examples illustrate correct examples of proper names ending a sentence. 
% The superposition of multiple function in a single neuron makes its interpretation challenging. Recently, SAEs have been shown to 

\vspace*{1.2cm}
\begin{figure*}[!htb]\centering
\sidesubfloat[]{\includegraphics[width=0.7\textwidth]{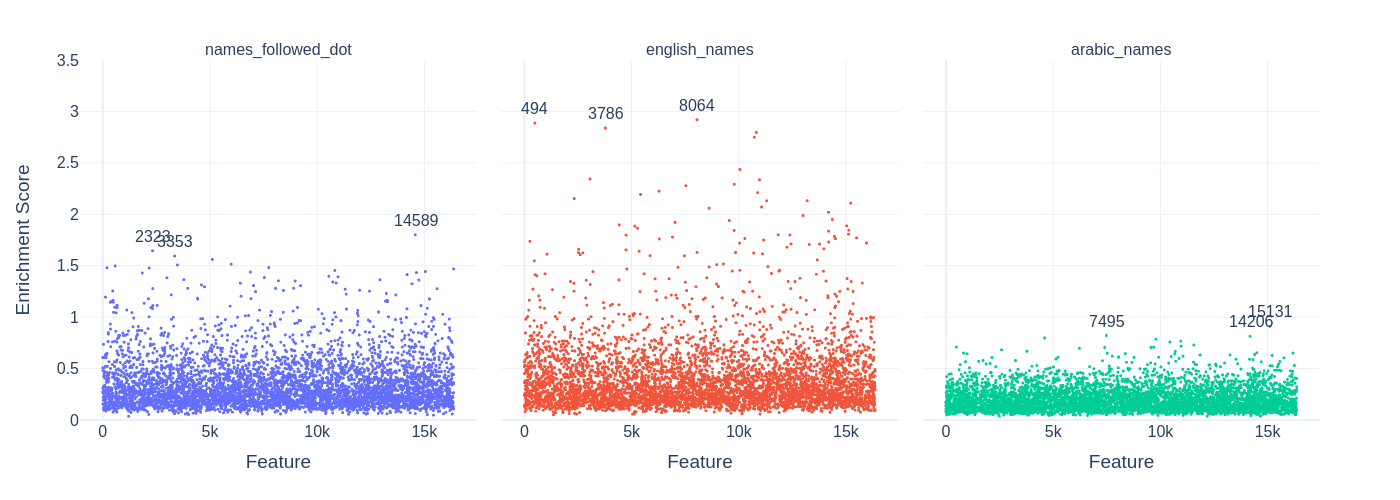}\label{fig:a}}
\\ %\hfil 
\sidesubfloat[]{\includegraphics[width=0.8\textwidth]{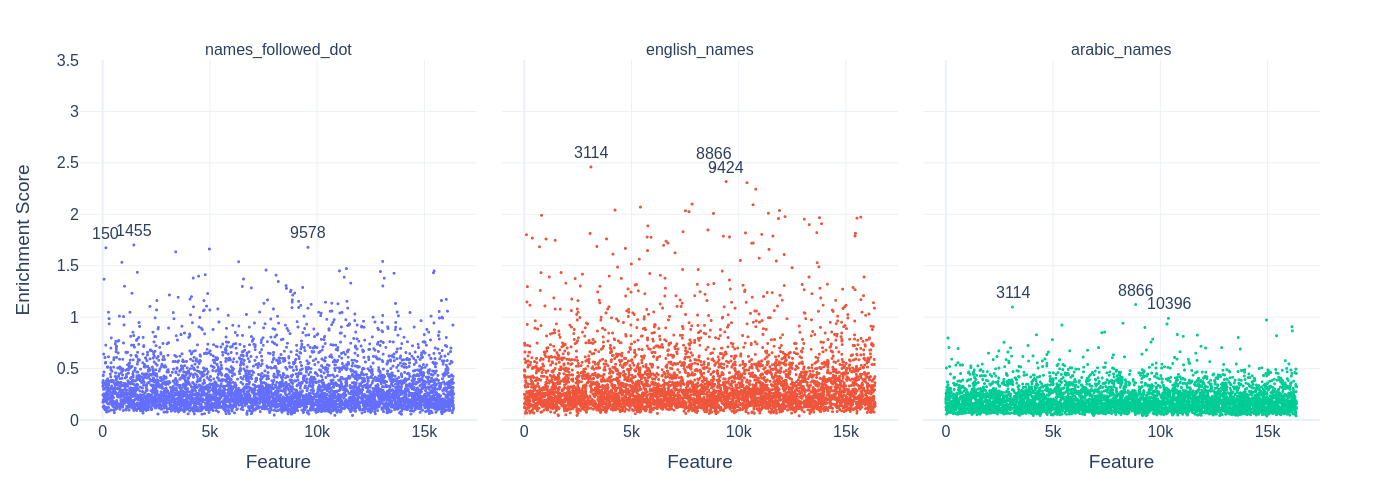}\label{fig:b}}
\caption{Manhattan plots of Enrichment Scores in (a) base and (b) further pre-trained models.}
\label{fig:manhattan}
\end{figure*}

% \begin{figure}
%     \centering
%     \includegraphics[width=1\linewidth]{scatter_names_dots_translation_data.png}
%     \caption{Enter Caption}
%     \label{fig:scatter_translation}  
% \end{figure}

% \begin{figure}
%     \centering
%     \includegraphics[width=0.5\linewidth]{scatter_names_dots_synthetic_data.png}
%     \caption{Enter Caption}
%     \label{fig:enter-label}
% \end{figure}
\section{Related Work}
\subsection{Continual pre-training}
Continual pre-training has been previously used for domain adaption in LMs~\cite{gupta2023continual, ke2023continual}. For example, general-purpose LLMs are continually pre-trained  for domain specific applications such as in financial applications~\cite{xie2023efficient} or new languages such as in AceGPT and SeaLLMs~\cite{huang2023acegpt, nguyen2023seallms}. 
\subsection{Mechanistic Intepretability}
Despite the progress in training capable language models, underlying working mechanisms are still not well understood. This presents risks and challenges in terms of model safety and robustness, as the model may produce inconsistent, irrelevant, or harmful outputs when faced with unfamiliar or adversarial inputs. The ability to quantify and understand working mechanisms in language models at different phases of model development is an important research topic. The field of mechanistic interpretability aims at providing principled tools and methods to reverse engineer working mechanisms in language models~\cite{elhage2021mathematical, elhage2022toy}.

% Example of coherent completion

\section{Conclusion}
In this work, we investigate the role of machine translated data in training language models for story generation in Arabic. We identify linguistic and cultural bias issues in the translated data. We propose to address these issues by further pre-training the models with a small amount of high-quality synthetic data. We investigate the effects of this intervention via Dictionary Learning tools. The trained Sparse Auto-Encoders shows a shift in the learned features towards corrected linguistic properties and reduced cultural bias. 

% \clearpage
\section{Limitations}
\subsection{Limitations of GPT-Eval} 
The use of LLM-as-judge approach to evaluate open-ended generation comes with some limitations. The complexity of the task of evaluating consistency, grammar and creativity is sometimes challenging for GPT-4. We noticed its limitations in complex Arabic evaluation compared to English. A reassessment of GPT-4 in Arabic with respect to newer models such as Llama-3 and Command R+ would be helpful to decide on the best LLM judge. 
\subsection{Generalization of the Results to Larger Datasets}
While TinyStories provides a testbed for exploring different facets of language models, the extension of this work to larger models faces challenges: as the model size grows, there is a need for a large dataset which shifts the burden on synthesizing diverse and high-quality dataset. The work done in models such as Phi-2 and Phi-3 are promising direction to address this challenge~\cite{abdin2024phi, li2023textbooks}.
\subsection{Extension to Other Domains}
Developing small generative language models that exhibits emergent properties is challenging for tasks beyond creative writing. For example, applications related to generic question answering would require large amount of data and larger models. Further research is needed to identify interesting applications and define the requirements in terms of data volume and quality to train highly capable small language models for other applications.
\subsection{Instruction Fine-Tuning}
This work is limited to the training of base models. An instruction fine-tuning dataset of English TinyStories is available and similar work in the paper could be applied for the analysis of fine-tuned models. We will investigate this in future work.

% You can obtain the complete ACL Anthology as a Bib\TeX{} file from \url{https://aclweb.org/anthology/anthology.bib.gz}.
% To include both the Anthology and your own .bib file, use the following instead of the above.

% Please see Section~\ref{sec:bibtex} for information on preparing Bib\TeX{} files.

\bibliography{latex/acl_latex}

% \newpage
\appendix
\renewcommand{\thefigure}{A.\arabic{figure}}
\setcounter{figure}{0}

\section{Story Samples}
\label{sec:appendix}
\subsection{Example NLLB-3B translation}
Here is an example of a story and its translated version using NLLB-3B.

\subsubsection{Original story in English} \textit{One day, a little boy named Tim went to the park.He saw a big tree and wanted to climb it.Tim was frightened, but he took a deep breath and started to climb. As he went up, he heard a small voice.The small voice said, "Please help me!"Tim looked around and saw a little bird.The bird had a hurt wing and could not fly.The bird was frightened too.Tim wanted to help the bird, so he thought of a plan.Tim took a big leaf from the tree and folded it.He made a soft bed for the bird. He put the bird on the leaf and climbed down.The bird was so happy and said, "Thank you, Tim!"The little boy and the bird became good friends.}
\\

\subsubsection{Translated story using NLLB-3B}
{\begin{arabtext} 
 \small 
 \arabtrue
 في يوم من الأيام ذهب صبي صغير يدعى تيم إلى الحديقة. رأى شجرة كبيرة وأراد أن يتسلقها. خاف تيم، ولكنه أخذ نفسا
 عميقا وبدأ في التسلق. عندما صعد، سمع صوتًا صغيراً. قال الصوت الصغيراً: `من فضلك ساعدني' نظر تيم حوله ورأى طيرًا صغيرًا. كان للطائر جناح مجروحًا ولم يستطع الطيران. كان الطير خائفًا أيضًا. أراد تيم مساعدة الطائر، لذا فكّر في خطة. أخذ تيم ورقة كبيرة من الشجرة وطويها. صنع للطائر سريرًا ناعمًا. وضع الطائر على الورقة و تسلق للأسفل. كان الطائر سعيد جدا وقال: `شكرا لك، تيم!' أصبح الصبي الصغير والطائر أصدقاء جيدين.
\end{arabtext} }
% \end{figure*}

\subsection{Synthetic generation using Command~R+}
\label{appx:synth}
We prompt Command R+ to synthesize a small set of high-quality Arabic stories as describe in~\ref{ssec:synth}. The prompt we used is shown in Figure~\ref{fig:prompt1}, and the story generation features in Table~\ref{tab:gen_features}. 

Here is an example of a generated Arabic story:
\\
    {\begin{arabtext}
\small 
كان هناك طفل صغير اسمه أحمد. أحمد كان مستعجل دائما ويحب اللعب في الخارج تحت الشمس. في أحد الأيام، بينما كان أحمد يلعب، بدأت السماء تمطر فجأة. أحمد لم يحب المطر أبدا لأنه كان يبلل ملابسه ويمنعه من اللعب. فكر أحمد في حل وقرر أن يصنع شمسية بنفسه حتى يستطيع اللعب في الخارج حتى عندما تمطر السماء.ذهب أحمد إلى أمه وسألها: `أمى، هل يمكن أن تساعديني في صنع شمسية؟ أريد أن ألعب في الخارج ولكن المطر يزعجني.'ابتسمت أمه وقالت: `بالتأكيد يا أحمد. يمكننا استخدام بعض الأقمشة القديمة وعصا طويلة لصنع شمسية رائعة. لكن يجب أن تكون صبورا لأن صنع الشيء يأخذ وقتا.'أحمد لم يكن صبورا أبدا لكنه وافق لأن أراد شمسية جميلة. عمل أحمد مع أمه بجد لعدة أيام. قطعوا القماش ولصقوه معا وحولوا العصا الطويلة إلى مقبض للشمسية. أخيرا، انتهوا من صنع الشمسية وكانت رائعة!أحمد كان سعيدا جدا وخرج إلى الخارج ليجرب شمسيته الجديدة. لكن، عندما فتحها، حدث شيء غير متوقع. الرياح بدأت تهب بقوة وأحمد لم يستطع السيطرة على الشمسية. طارت الشمسية من يده واختفت في السماء. أحمد بكى وحزن جدا لأنه فقد شمسيته الجميلة التي عمل بجد لصنعها.الأم جاءت لعنده وحاولت أن تواسيه: `لا تبكي يا أحمد. أنا فخورة بك لأنك عملت بجد وصنعت شيئا رائعا. لكن تذكر، عندما نريد شيئا ما، يجب أن نكون صبورين ونعمل بجد لتحقيقه. في بعض الأحيان، حتى إذا فعلنا كل شيء بشكل صحيح، قد لا نحصل على ما نريد. هذا جزء من الحياة ويجب أن نتعلم كيف نتعامل مع خيبات الأمل.أحمد تعلم درسا مهما ذلك اليوم. تعلم أن العمل الجاد والصبر مهمان، ولكن أيضا أن هناك أشياء خارجة عن سيطرتنا. في بعض الأحيان، حتى إذا احتجنا شيئا ما بشدة، قد لا نحصل عليه. هذا لا يعني أن نتوقف عن المحاولة، ولكن أن نتعلم كيف نتأقلم ونستمر في النمو.  
\end{arabtext} }

% \subsection{Evaluation Prompt for GPT-Eval}
 \begin{figure}[h!]
    \centering
    \begin{tikzpicture}
        \node[draw, rounded corners, align=center] {
            \begin{minipage}{0.95\columnwidth}
                    \texttt{Write a short story  (3-5 paragraphs) in Arabic language which only uses very simple words that a 3 year old child would likely understand. The story should use the verb "\{verb\}", the noun "\{noun\}" and the adjective "\{adjective\}". The story should have the following features: \{features\}. Remember to write in Arabic and to only use simple words!} % Set the text in typewriter font
            \end{minipage}
        };
    \end{tikzpicture}
    \caption{Prompt for synthesizing Arabic stories with Command R+, an Arabic-capable LLM. }
    \label{fig:prompt1}
\end{figure}

\begin{table*}
\begin{tabular}{p{0.21\textwidth}|p{0.04\textwidth}|p{0.55\textwidth}}
\toprule
Feature & Prob. & Instruction \\
\midrule
Dialogue & 0.6 & the story should contain at least one dialogue \\
BadEnding & 0.3 & the story has a bad ending \\
Conflict & 0.1 & the story has some form of conflict in it \\
Moral Value & 0.1 & the story has a moral value \\
Fore-shadowing & 0.1 & the narrative uses foreshadowing or setup and payoff \\
Twist & 0.3 & something unexpected happens / there is a plot twist \\
\bottomrule
\end{tabular}
\caption{For each story, a set of features is randomly selected  according to the probability and definition described in the table.}
\label{tab:gen_features}
\end{table*}

\subsection{Completion example}
Here is an example of a prompt from the test set (top panel of Figure~\ref{fig:dd}) and model 33M-ar completion (bottom panel).
\begin{figure}[h!]
\centering
\begin{tikzpicture}
\node[draw, rounded corners, align=center] {
\begin{minipage}{0.95\columnwidth}
\begin{arabtext}
\small 
في يوم من الأيام، كانت الجدة في المطبخ تستعد للطهي. لديها مكونات لطبق شهي. قالت الجدة لحفيدتها: `هل ستساعديني في إعداد هذه الوجبة؟' ردت حفيدتها، جيني: `نعم، سأساعدك!' بدأت الجدة وجيني في تحضير المكونات. كانت جيني غير ماهرة. أسقطت ***
\end{arabtext}
\end{minipage}
};
\end{tikzpicture}
\begin{tikzpicture}
\node[draw, rounded corners, align=center] {
\noindent\fcolorbox{white}{chartreuse}{%
\begin{minipage}{0.92\columnwidth}
\begin{arabtext}
\small 
  الدقيق في الفرن بينما كانوا يعدون العشاء على الموقد. كان رائحة المطبخ لذيذة. عندما كان العشاء جاهزاً ، وضعت الجدة وجيني وعاءً كبيرًا من الطعام اللذيذ على الطبق. بدا لذيذًا جدًا. أعطت الجدة جيني ملعقة كبيرة لتحريك الطعام...
\end{arabtext}
\end{minipage}
}};
\end{tikzpicture}
\caption{An example of a completion by the model 33M-ar. The top panel shows the prompt from the test set ending in an incomplete sentence followed by *** to assess the model's ability to complete a coherent transition sentence. The bottom panel shows the model continuation.}
    \label{fig:dd}
\end{figure}

\section{Training Details}
\subsection{Process calibration using original TinyStories}
\label{appx:calibration}
\begin{figure}
    \centering
    \includegraphics[width=0.93\linewidth]{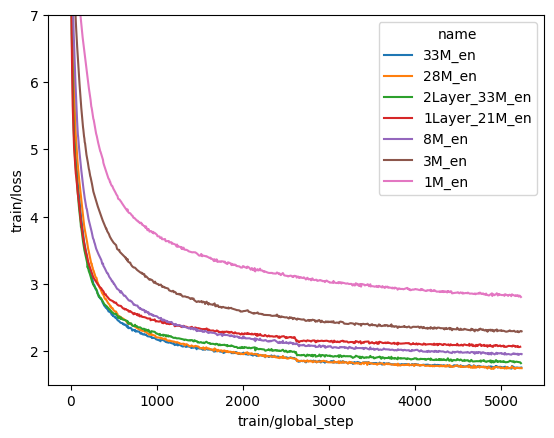}
    \caption{Train loss of  models trained on  English TinyStories.}
    \label{fig:loss_en}
\end{figure}

We calibrate the training process by  first training the chosen  architectures  on the original TinyStories dataset and comparing the benchmarks with the published results. Table~\ref{tab:model_en} shows GPT-4 scores for some trained models, and Figure~\ref{fig:loss_en} shows the loss curve for the training.

\begin{table}[ht]
\centering
\begin{tabular}{lccc} \toprule
\small{model} & \small{Grammar} & \small{Creativity} & \small{Consistency} \\ \midrule
\small \textbf{gpt2-large} & 5.52 & 1.31 & 1.68 \\
\small \textbf{33M-en} & 6.40 & 6.63 & 6.97 \\ %\midrule
\small \textbf{Mistral-7b} & 7.86 & 6.88 & 8.04 \\ 
\bottomrule
\end{tabular}
\caption{Performance results of a model trained on the original TinyStories dataset against baselines (gpt2-large~\cite{radford2019language} and Mistral-7B~\cite{jiang2023mistral}) in order to calibrate the training process.}
\label{tab:model_en}
\end{table}

\subsection{Pre-training loss curves}
\label{appx:pt_loss}
\begin{figure}
    \centering
    \includegraphics[width=1.04\linewidth]{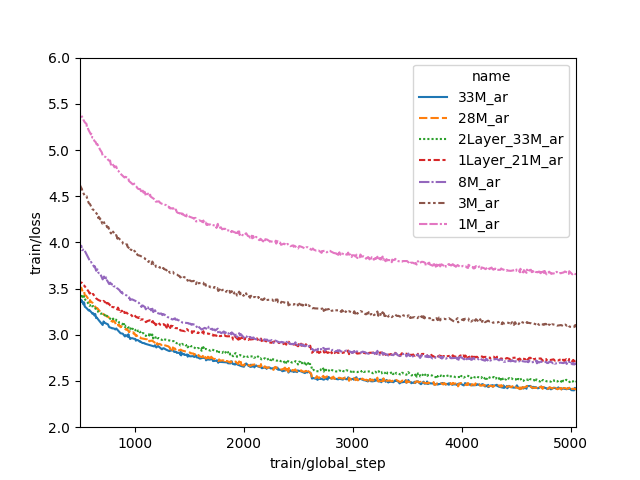}
    \caption{Training loss of models trained on translated TinyStories.}
    \label{fig:loss_train_ar}
\end{figure}

\begin{figure}[h!]
    \centering
    \includegraphics[width=1.07\linewidth]{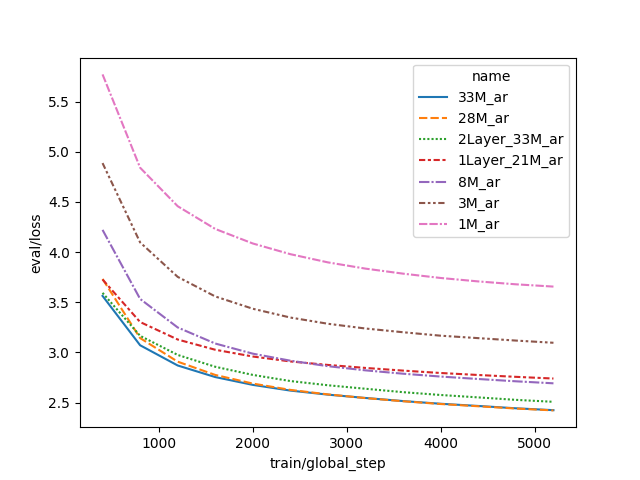}
    \caption{Validation loss of models trained on translated TinyStories.}
    \label{fig:loss_eval_ar}
\end{figure}

Figures~\ref{fig:loss_train_ar} and \ref{fig:loss_eval_ar} show the training and validation loss, respectively, for the pre-training stage using translated TinyStories.

\subsection{Continual pre-training loss curves}
\label{appx:cpt_loss}
\begin{figure}[h!]
    \centering
    \includegraphics[width=1\linewidth]{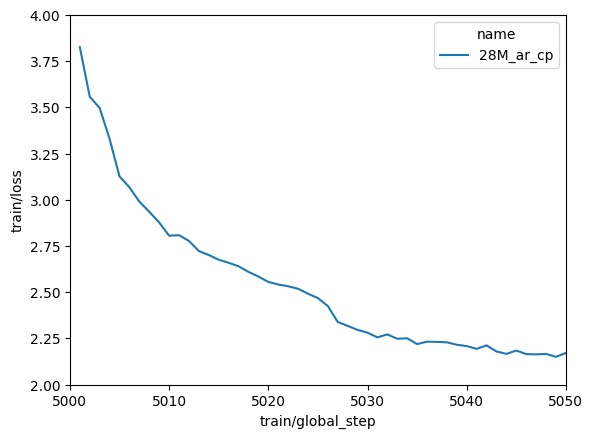}
    \caption{Training loss during continual pre-training with high-quality synthesized data.}
    \label{fig:loss_train_ar_fpt}
\end{figure}

\begin{figure}[h!]
    \centering
\includegraphics[width=1\linewidth]{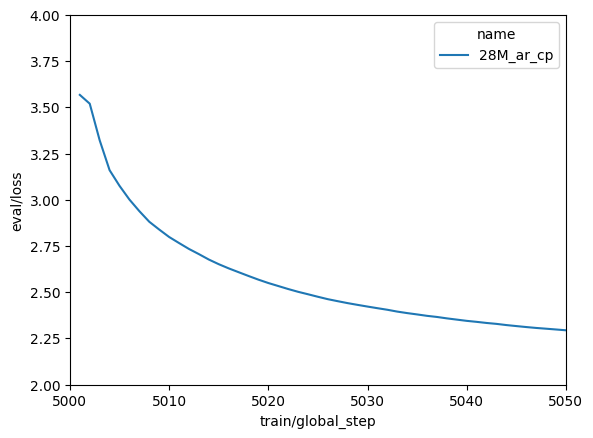}
    \caption{Evaluation loss during continual pre-training with high-quality synthesized data.}
    \label{fig:loss_eval_ar_fpt}
\end{figure}

Figures~\ref{fig:loss_train_ar_fpt} and \ref{fig:loss_eval_ar_fpt} show the training and validation loss, respectively, for the continual pre-training stage using the Arabic synthetic story dataset.

\subsection{Feature dashboard examples}
\begin{figure*}
    \centering
    \includegraphics[width=1\linewidth]{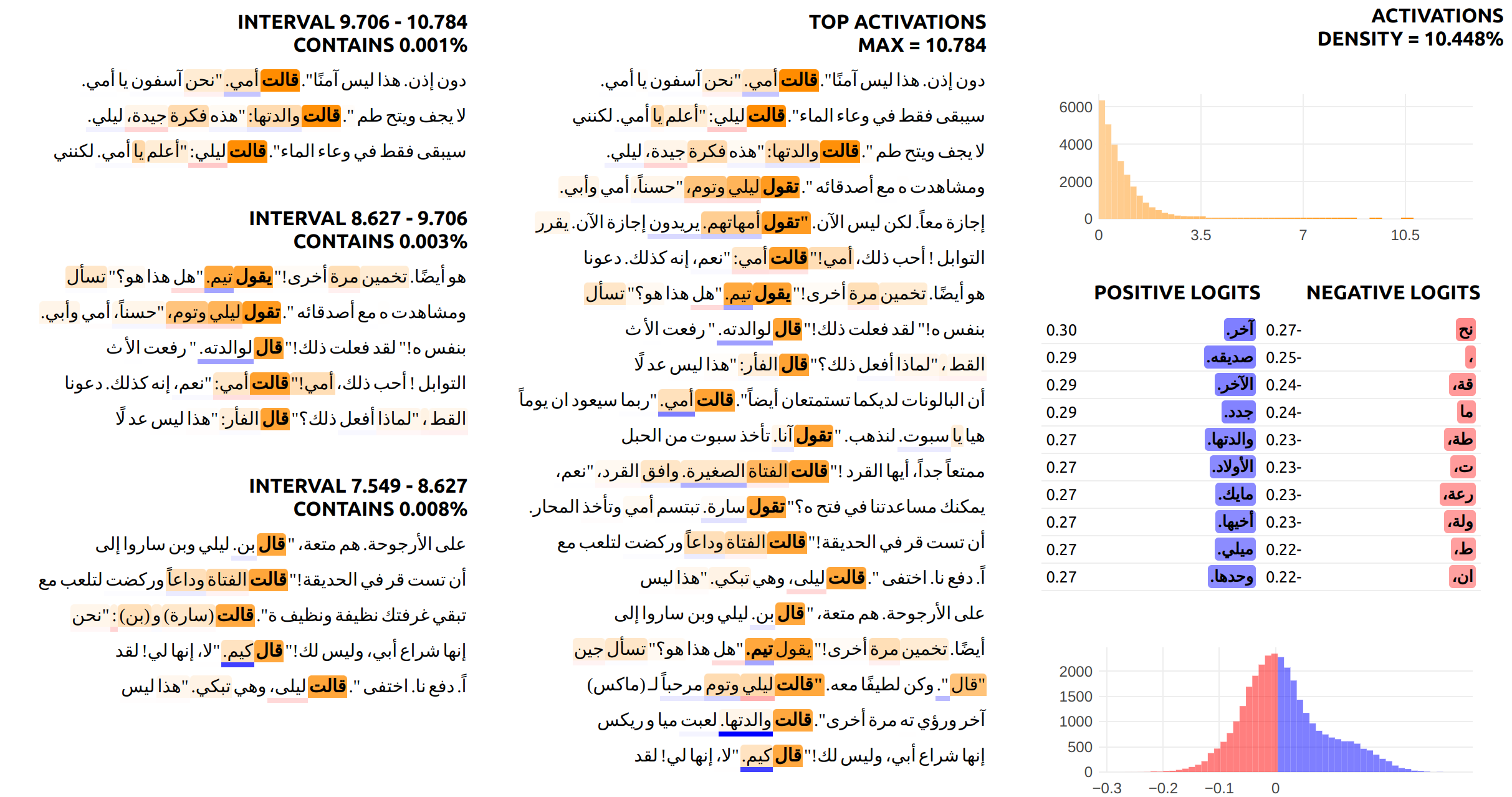}
    \caption{Dashboard of feature \#14589 from SAEs trained on the last MLP layer of the base model 2L-33M-ar. It corresponds to token set in TSEA formed as first name followed by punctuation. This feature captures the quote tagging issue discussed in Section~\ref{sec:mt_issues}.}
    \label{fig:dashb1}
\end{figure*}

Figure \ref{fig:dashb1}  is a dashboard summarizing the properties of the selected feature \#14589. The top right histogram gives the activation distribution for the feature. The panel underneath shows the highest and lowest ranked tokens based on the logit difference when the feature is ablated. The text panels show training samples for different feature activation intervals. Feature \#14589 captures the verb 'say' in Arabic in different tense forms following a quote. This construction is not correct in Arabic as the use of the quote tagging is given before the quote. This reversed structure has been inherited from English due to the low-quality translation of the dataset. This feature helped us reveal some of the limitations of our model trained with low-quality translation data. 
% Feature \ref{fig:dashb2} illustrates a learned property by the continually trained model. 


% \begin{figure*}
%     \centering
%     \includegraphics[width=1\linewidth]{feat150_synthetic_data.png}
%     \caption{Dashboard of feature \#150 from SAEs trained on the last MLP layer of continually pre-trained model. It corresponds to token set in TSEA formed as first name followed by punctuation. This feature activates on correct quote tagging format. The further pre-training model learned to fix the quote tagging issue inherited from inaccurate translation.}
%     \label{fig:dashb2}
% \end{figure*}

% \begin{figure}
%     \centering
%     \includegraphics[width=1\linewidth]{activation.png}
%     \caption{feature 37}
%     \label{fig:enter-label}
% \end{figure}

% \begin{figure}
%     \centering
%     \includegraphics[width=0.9\linewidth]{top_activation.png}
%     \caption{feature 37}
%     \label{fig:enter-label}
% \end{figure}

\end{document}